\documentclass{article}

\PassOptionsToPackage{table}{xcolor}
\usepackage[final]{corl_2026}

\title{Pre-trained Image Generators for\\Adaptive Robot Simulation}

\author{
  Jiong Lin, Yihong Liu, Judah Goldfeder, Hod Lipson\\[2pt]
  {\normalfont Columbia University}
}

\usepackage{booktabs}
\usepackage{multirow}
\usepackage{graphicx}
\usepackage{svg}
\usepackage{placeins}

\usepackage[hypcap=false,compatibility=false]{caption}

\usepackage{array}
\usepackage{gensymb}
\usepackage{fancyvrb}
\usepackage{lipsum}
\usepackage{xspace}
\usepackage{pifont}
\usepackage{cuted}
\usepackage{amssymb}
\usepackage{amsmath}
\usepackage{algorithmicx}

\makeatletter
\renewcommand{\paragraph}{\@startsection{paragraph}{4}{\z@}{0.4ex \@plus 0.2ex \@minus 0.1ex}{-1em}{\normalsize\bf}}
\makeatother

\definecolor{rankone}{HTML}{9EC5EA}
\definecolor{ranktwo}{HTML}{C9DEF4}
\definecolor{rankthree}{HTML}{E0ECF8}
\newcommand{\shadekey}{Shading: \colorbox{rankone}{best}, \colorbox{ranktwo}{second}, \colorbox{rankthree}{third} in Table~\ref{tab:main}; the other tables shade only the best.}

\definecolor{actfill}{HTML}{DCEFD7}\definecolor{actline}{HTML}{6AAE5E}
\definecolor{camfill}{HTML}{F8E2D2}\definecolor{camline}{HTML}{D69A72}
\definecolor{modfill}{HTML}{D6E4F5}\definecolor{modline}{HTML}{5B8FCF}

\usepackage{amsfonts}
\usepackage{bm}
\usepackage{wrapfig}
\usepackage{float}
\usepackage{enumitem}
\usepackage{algorithm}
\usepackage{algpseudocode}
\usepackage{afterpage}

\makeatletter
\renewcommand{\@noticestring}{CoRL 2026 Workshop on The Action Gap (TAG 2026), Austin TX, USA.}
\makeatother
\hypersetup{
  pdftitle={Pre-trained Image Generators for Adaptive Robot Simulation},
  pdfauthor={Jiong Lin, Yihong Liu, Judah Goldfeder, Hod Lipson},
  pdfsubject={CoRL 2026 Workshop on The Action Gap (TAG 2026)}
}

\definecolor{act}{HTML}{6AAE5E}
\definecolor{cam}{HTML}{D69A72}
\newcommand{\av}{\bm{a}}
\newcommand{\cv}{\bm{c}}

\newcommand{\fire}{\raisebox{-0.12em}{\includegraphics[height=0.95em]{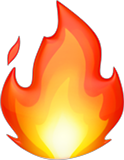}}}
\newcommand{\snow}{\raisebox{-0.12em}{\includegraphics[height=0.95em]{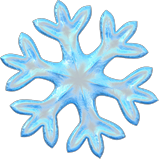}}}

\begin{document}
\maketitle

\begin{abstract}
Action-conditioned generative models of robots usually use egocentric views from the robot's cameras, which limit the view to the gripper and its workspace and leave the rest of the body unseen.
We study the third-person view: given one image of a robot with its end-effector and camera poses, render the whole robot at a new end-effector pose, from a new camera pose, or both.
At test time, the model needs no URDF, joint angles or simulator.
We fine-tune the image generator FLUX.2 with LoRA on image pairs streamed from MuJoCo, and give it the end-effector and camera poses as condition tokens together with a per-pixel end-effector ray heatmap.
On a 5-DoF WidowX~200 in two render styles and a 7-DoF Franka FR3, the predicted arm reaches a body IoU, the mean IoU of the arm links below the wrist, of $0.27$--$0.53$, against at most $0.08$ for nearest-neighbor retrieval and $0.19$ for large language models that write rendering code.
When the robot's upper arm is bent, continuing to train the model on 30--300 trajectories of the modified robot adapts it toward the new body.
\end{abstract}

\begin{figure}[h]
  \centering
  \includegraphics{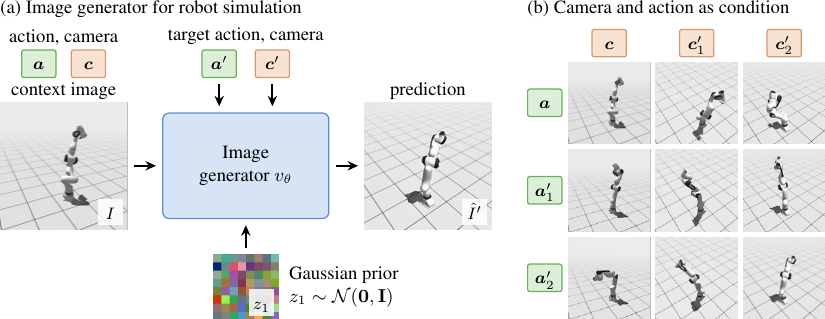}
  \caption{\textbf{Third-person robot simulation with an image generator.} (a)~From a context image with its action (end-effector pose) and camera pose, a target action and camera pose, and a sample of Gaussian noise, the image generator predicts the target image. (b)~Eight predictions of one model from one input image (top left) under new actions (rows) and new cameras (columns); realistic Franka FR3, first test scene.}
  \label{fig:teaser}
\end{figure}

\section{Introduction}
\label{sec:intro}

Multiple 2D views of a scene from known cameras constrain its 3D geometry~\cite{hartley2004mvg}.
For example, structure from motion recovers a scene from images alone~\cite{sfm}, and novel-view synthesis renders a scene from new cameras~\cite{nerf,3dgs}.
World models~\cite{worldmodels,dreamer} also learn from recorded actions and predict what follows each action, often in a learned latent space.
Their training data range from large video collections~\cite{genie} to recorded human play of Minecraft~\cite{dreamer4} and recorded agent play of DOOM~\cite{gamengen}.

In robotics, world models are usually action-conditioned video generators that act as learned simulators~\cite{unisim}.
Most of them use egocentric views, which mainly show the gripper and its workspace~\cite{evac}.
In contrast, robots that learn models of their own bodies, known as self-models~\cite{bongard2006resilient}, represent the whole body~\cite{chen2022fullbody}.
A whole-body view matters for obstacle avoidance, mobile robots and multi-robot workspaces.
We therefore study a third-person setting (Fig.~\ref{fig:teaser}).
Given one image of a robot with its end-effector and camera poses, the model renders the robot at a target end-effector pose from a target camera pose.
Without a URDF, joint angles or simulator at test time, the model infers an arm configuration that reaches the target pose, as inverse kinematics would.
The task thus adds the action to single-image novel-view synthesis~\cite{zero123}, which changes only the camera.

We address this task by fine-tuning the pre-trained image generator FLUX.2 Klein~\cite{flux2klein} with LoRA~\cite{lora} on image pairs that MuJoCo~\cite{mujoco} renders online.
The action enters as condition tokens and a per-pixel \emph{end-effector ray heatmap}, the distance from each pixel's camera ray to the target end-effector position.
Our contributions are:
(1)~a third-person, whole-body formulation of action-conditioned image generation, scored by the mean IoU of the arm links below the wrist (body IoU);
(2)~a model with a pixel-aligned action input reaching body IoU $0.27$--$0.53$, against at most $0.08$ for retrieval from 25M simulated frames and $0.19$ for frontier language models writing rendering code;
and (3)~evidence that continued training adapts the model toward a new body: shown a bent arm, the model first draws its original arm, and 30--300 trajectories of the bent robot adapt it with over $30\times$ less compute than training anew.

\section{Related Work}
\label{sec:related}

\paragraph{Image generators.}
Diffusion models generate images by iterative denoising~\cite{ddpm}, with a U-Net in an autoencoder's latent space~\cite{ldm}.
Flow matching and rectified flow instead learn a velocity field carrying noise to data along near-straight paths~\cite{lipman2023flow,liu2023rectified}, and current image generators such as FLUX.2 Klein~\cite{flux2klein} pair this objective with a transformer and also edit images from text instructions~\cite{kontext,flux2klein}.
Camera-controlled generators synthesise new views from one image, conditioned on the relative camera pose~\cite{zero123} or per-pixel ray maps~\cite{cameractrl,cat3d}, and 4DiM controls space and time together~\cite{fourdim}.
Building on these generators, we fine-tune FLUX.2 Klein with LoRA~\cite{lora} and give it two pixel-aligned inputs: Pl\"ucker rays for the camera and a ray heatmap for the action.

\paragraph{Generative models in robotics.}
Video generators conditioned on robot actions act as learned simulators~\cite{unisim,irasim,cosmos}; for example, Ctrl-World~\cite{ctrlworld} evaluates policies in imagination.
EnerVerse-AC~\cite{evac} encodes the end-effector pose as a map projected into the robot's egocentric cameras, with ray maps for the camera.
Closer to our setting, Robot-Factored World Models~\cite{robotfactored} render the full robot through a URDF, and MVISTA-4D~\cite{mvista4d} generates multi-view futures from an encoded action trajectory.
Image editing models have also been proposed as world models~\cite{sweet} and as backbones of world action models~\cite{imagewam}.
Like them, we build on an image generator, here to represent the robot's whole body in image space.

\paragraph{World models and self-models.}
World models predict how an environment responds to an agent's actions~\cite{worldmodels}, in image space or in a learned representation space~\cite{jepa}.
These models learn games from videos and game engines~\cite{genie,gamengen,diamond} and navigation from recorded trajectories~\cite{nwm}, and serve as environments for training agents~\cite{dreamer4,dreamer}.
Learned simulators also distil analytical ones for particles~\cite{gns} and robot dynamics in state space~\cite{nerd}.
Like Melissa~\cite{melissa}, we train on data streamed online from the simulator.
Self-models are robots' learned models of their own bodies~\cite{bongard2006resilient,kwiatkowski2019task}: signed distance fields conditioned on joint angles~\cite{chen2022fullbody}, neural fields and articulated Gaussian splats~\cite{hu2025teaching,hu2025gs,drrobot}, Jacobian fields from video~\cite{njf}, kinematic structure from point clouds~\cite{autourdf}, and flow-matching models of articulated mechanisms and continuum robots~\cite{articflow,tdcr}.
Self-models support adaptation to damage~\cite{bongard2006resilient}, motivating our bent-arm study.
Our model joins both lines of work: an action-conditioned image generator of the robot's whole body.

\section{Method}
\label{sec:method}

\begin{figure}[t]
  \centering
  \includegraphics{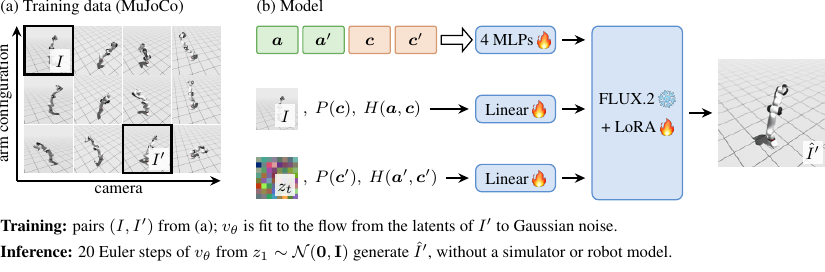}
  \caption{\textbf{Method.} (a)~Inside the training loop, MuJoCo renders a trajectory of arm configurations (rows) from four cameras (columns), and a training pair $(I, I')$ is drawn from it. (b)~Each latent token of the context image ($\mathcal{E}$: frozen VAE encoder) and of the noisy target $z_t$ is concatenated with the Pl\"ucker coordinates $P$ of its camera rays and its end-effector ray heatmap $H$, and embedded by FLUX.2's patch embedding, extended with trained weights for $P$ and $H$. \fire~trained, \snow~frozen.}
  \label{fig:method}
\end{figure}

\paragraph{Problem setting.}
A context image $I$ shows the robot with its end-effector at pose $\av \in SE(3)$, taken by a camera with pose $\cv \in SE(3)$ and known intrinsics.
Given a target end-effector pose $\av'$ and camera pose $\cv'$, the model predicts the image $\hat I'$ of the robot at $\av'$ seen from $\cv'$.
The action is an absolute end-effector pose in the base frame (position and 6D rotation~\cite{zhou2019continuity}); the gripper opening is fixed.
We evaluate three modes: \emph{new action} ($\av' \neq \av$, $\cv' = \cv$), \emph{new view} ($\av' = \av$, $\cv' \neq \cv$), and \emph{both}.

\begin{wrapfigure}{r}{0.5\textwidth}
  \vspace{-1.2em}
  \begin{minipage}{\linewidth}
  \hrule height 0.8pt \vspace{2pt}
  \captionof{algorithm}{Training with the online data engine}
  \label{alg:train}
  \vspace{-4pt}\hrule\vspace{2pt}
  \footnotesize
  \begin{algorithmic}
    \Statex \textit{Render workers, 20 per GPU, in parallel:}
    \Loop
      \State $q_1, q_2, q_3 \sim$ joint limits; interpolate $q_{1:8}$
      \State Resample on self-collision or floor contact
      \State $c_{1:4} \sim$ cameras around the robot
      \State Store $\mathcal{E}(\mathrm{render}(q_i, c_j))$ for all $i, j$
    \EndLoop
    \Statex \textit{Trainer:}
    \Loop
      \State $(I, \av, \cv), (I', \av', \cv') \sim$ buffer
      \State $z_0$: latent of $I'$; $z_1 \sim \mathcal{N}(\mathbf{0}, \mathbf{I})$; $t \sim$ logit-normal
      \State $z_t \gets (1-t)\,z_0 + t\,z_1$
      \State $P, H$ from $(\av, \cv)$ and $(\av', \cv')$ \Comment{Eq.~\ref{eq:embed}}
      \State $\theta \gets \theta - \eta\, \nabla_\theta \mathcal{L}$ \Comment{Eq.~\ref{eq:loss}}
    \EndLoop
  \end{algorithmic}
  \vspace{-2pt}\hrule height 0.8pt
  \end{minipage}
  \vspace{-1.0em}
\end{wrapfigure}

\paragraph{Data engine.}
MuJoCo~\cite{mujoco} renders the training data inside the training loop (Fig.~\ref{fig:method}a, Algorithm~\ref{alg:train}).
A \emph{scene unit} is a trajectory of 8 arm configurations rendered from 4 cameras.
The trajectory interpolates three random joint configurations within the joint limits, and every frame is checked for self-penetration and floor contact.
The four cameras surround the robot at random azimuths and elevations, at a distance that frames the whole robot.
Render workers on every GPU fill a buffer of 3{,}000 VAE-encoded units, from which training pairs are drawn: 30\% new action, 20\% new view and 50\% both.
Each unit serves about 7 pairs (20 for the colored WX200) before it is replaced, and test units come from disjoint random seeds.

\paragraph{Model.}
We fine-tune FLUX.2 Klein base 4B~\cite{flux2klein}, a rectified-flow transformer~\cite{lipman2023flow,liu2023rectified}, as shown in Fig.~\ref{fig:method}b.
Its VAE encoder turns a $128\times128$ image into an $8\times8$ grid of latent tokens, each covering $16\times16$ pixels.
As in FLUX.1 Kontext~\cite{kontext}, the context image enters as clean latent tokens next to the noisy target tokens; context tokens share the target tokens' positions with an offset of 10 on the first rotary axis.
Each latent token also receives two pixel-aligned signals for the 4 rays through the centres of its $2\times2$ sub-blocks: their Pl\"ucker coordinates $(\bm{d}, \bm{m})$ with $\bm{m} = \bm{o}\times\bm{d}$~\cite{cameractrl} ($P \in \mathbb{R}^{24}$), and the \emph{end-effector ray heatmap} $\exp(-(\delta/\sigma)^2)$, where $\delta = \lVert \bm{p}\times\bm{d} - \bm{m} \rVert$ is the ray's distance to the 3D end-effector position $\bm{p}$ and $\sigma \in \{3, 10\}$\,cm ($H \in \mathbb{R}^{8}$).
Context tokens use the context end-effector and camera poses, and target tokens the target ones.
For a latent token $\bm{x}_i$ of either image, the concatenation $[P_i; H_i]$ passes through one bias-free, zero-initialised linear layer $W_{PH}$ and is added to FLUX.2's patch embedding $W_{\mathrm{patch}}$:
\begin{equation}
  \bm{h}_i = W_{\mathrm{patch}}\, \bm{x}_i + W_{PH}\, [P_i; H_i].
  \label{eq:embed}
\end{equation}
The text encoder is replaced by four condition tokens, one for each of $\av, \av', \cv, \cv' \in \mathbb{R}^9$ (end-effector position and the first two rotation columns; camera eye, forward and up), and four learned register tokens.
Each condition is expanded with sine and cosine features at frequencies $2^k\pi$, $k=0,\dots,7$, and mapped to its token by its own two-layer MLP.
The transformer stays frozen, and LoRA adapters~\cite{lora} ($r = \alpha = 64$) on every linear layer of its 5 double-stream and 20 single-stream blocks add 96.3M trainable parameters; the new input layers add 32.3M.

\paragraph{Training and sampling.}
The model is a velocity field $v_\theta(z_t, t \mid I, \av, \cv, \av', \cv')$ over the target latents $z_t = (1-t)\,z_0 + t\,z_1$, which interpolate between the latents $z_0$ of $I'$ and Gaussian noise $z_1$.
We train it with the flow-matching loss
\begin{equation}
  \mathcal{L}(\theta) = \mathbb{E}_{t,\, z_0,\, z_1} \big\lVert v_\theta(z_t, t \mid I, \av, \cv, \av', \cv') - (z_1 - z_0) \big\rVert^2,
  \label{eq:loss}
\end{equation}
with $t$ drawn from a logit-normal distribution ($t = \mathrm{sigmoid}(n)$, $n \sim \mathcal{N}(0,1)$), AdamW ($\beta = (0.9, 0.99)$, no weight decay, gradient clipping at 1.0), learning rates of $10^{-4}$ for LoRA and $3\times10^{-4}$ for the new layers, 500 warm-up steps and a linear decay to 10\% over the final 2 hours.
Target-side condition tokens are dropped with probability 0.1 for classifier-free guidance~\cite{ho2022cfg}.
Sampling integrates $v_\theta$ from $z_1 \sim \mathcal{N}(\mathbf{0}, \mathbf{I})$ with 20 Euler steps and guidance scale 2.

\paragraph{Compute.}
\label{sec:compute}
Each model trains on 8 RTX~5090 GPUs at batch size 256: 11.0\,h on the colored WX200 (30.8k steps, 7.9M pairs), 6.0\,h on the realistic WX200 (16.0k steps, 4.1M pairs) and 7.0\,h on the realistic FR3 (18.65k steps, 4.8M pairs).
Adapting the realistic WX200 model to a bent arm (Sec.~\ref{sec:bent}) passes $0.36$ link IoU on the bent robot within 250 steps, about 6 minutes, while the same model trained from scratch on the bent robot reaches $0.24$ after 3 hours.
Sampling takes 0.31\,s per image at batch size 32.

\section{Experiments}
\label{sec:experiments}

\paragraph{Settings.}
We train one model per setting: a 5-DoF Trossen WidowX~200 (WX200) with flat per-link colors, the same robot with realistic materials and shadows, and a realistic 7-DoF Franka FR3, which is kinematically redundant.
Each setting has a test set of 192 pairs (64 per mode) rendered separately from the training stream, and we report the final checkpoint of each run.
All methods are compared on the first 16 pairs of each mode (48 per setting), since the language-model baselines take minutes per query; Appendix~\ref{app:permode} gives all 192 pairs.

\paragraph{Metrics.}
We label every image per link: by palette for the colored robot, and for the realistic robots by MuJoCo segmentation for the ground truth and a U-Net segmenter trained in simulation for predictions.
\emph{Body IoU} is the mean per-link IoU over the arm links below the wrist (WX200 shoulder to lower forearm, FR3 links 1--5), and \emph{link IoU} the mean over all links; \emph{gripper error} is the pixel distance between predicted and true gripper centroids.
Tables give mean $\pm$ std over test pairs; comparisons of means use 95\% bootstrap intervals.

\paragraph{Baselines.}
\emph{Copy input} returns the context image, and \emph{random frame} a random training frame.
\emph{Nearest neighbor} (NN) returns the frame whose end-effector and camera poses are closest to the target among 1k or 25M frames from the data engine; 25M is about the number of frames each model trains on.
\emph{FLUX.2 (zero-shot)} uses FLUX.2 Klein without fine-tuning as an image editor, prompted with the query as text.
Two frontier language models, Claude Opus~5.5 and GPT-6 Astra, receive the context image, both end-effector and camera poses, both end-effector pixel positions and coordinate conventions, but not the robot's name, and write a Python program that renders the target image in a sandbox (Appendix~\ref{app:zeroshot}).
The \emph{VAE ceiling} encodes and decodes the ground truth.

\subsection{Main results}
\label{sec:main}

\begin{table}[t]
  \centering
  \caption{\textbf{Main results} on 48 test pairs per setting (16 per mode), mean $\pm$ std over pairs; ours: 4 sampling seeds. Body: mean IoU of the arm links not placed directly by the end-effector input. Grip: gripper centroid error in pixels ($128\times128$).}
  \label{tab:main}
  \footnotesize
  \setlength{\tabcolsep}{2pt}
  \begin{tabular*}{\textwidth}{@{\extracolsep{\fill}}lcccccc}
\toprule
 & \multicolumn{2}{c}{Colored WX200} & \multicolumn{2}{c}{Realistic WX200} & \multicolumn{2}{c}{Realistic FR3} \\
\cmidrule(lr){2-3}\cmidrule(lr){4-5}\cmidrule(lr){6-7}
Method & Body$\uparrow$ & Grip$\downarrow$ & Body$\uparrow$ & Grip$\downarrow$ & Body$\uparrow$ & Grip$\downarrow$ \\
\midrule
Random frame & 0.021$\pm$0.030 & 35.6$\pm$16.6 & 0.017$\pm$0.041 & 38.0$\pm$17.4 & 0.037$\pm$0.045 & 46.5$\pm$21.4 \\
Copy input & 0.092$\pm$0.106 & 32.3$\pm$20.1 & \cellcolor{rankthree}0.103$\pm$0.158 & 35.2$\pm$21.4 & \cellcolor{ranktwo}0.119$\pm$0.151 & 46.1$\pm$20.0 \\
NN (1k) & 0.032$\pm$0.043 & 24.3$\pm$12.2 & 0.028$\pm$0.042 & 22.8$\pm$15.3 & 0.033$\pm$0.050 & 29.0$\pm$16.6 \\
NN (25M) & 0.053$\pm$0.079 & 12.4$\pm$7.4 & 0.025$\pm$0.035 & \cellcolor{rankthree}13.7$\pm$7.6 & 0.079$\pm$0.095 & \cellcolor{ranktwo}17.3$\pm$16.4 \\
FLUX.2 (zero-shot) & 0.065$\pm$0.079 & 32.4$\pm$21.0 & 0.061$\pm$0.083 & 35.5$\pm$17.7 & 0.048$\pm$0.105 & 41.1$\pm$17.9 \\
Claude Opus 5.5 & \cellcolor{rankthree}0.161$\pm$0.098 & \cellcolor{rankthree}8.3$\pm$9.0 & 0.073$\pm$0.123 & 20.4$\pm$19.9 & 0.080$\pm$0.068 & 35.2$\pm$19.9 \\
GPT-6 Astra & \cellcolor{ranktwo}0.187$\pm$0.156 & \cellcolor{ranktwo}6.6$\pm$3.6 & \cellcolor{ranktwo}0.165$\pm$0.147 & \cellcolor{ranktwo}7.0$\pm$7.7 & \cellcolor{rankthree}0.116$\pm$0.139 & \cellcolor{rankthree}20.5$\pm$14.5 \\
Ours & \cellcolor{rankone}0.531$\pm$0.190 & \cellcolor{rankone}2.2$\pm$2.2 & \cellcolor{rankone}0.346$\pm$0.219 & \cellcolor{rankone}1.3$\pm$1.4 & \cellcolor{rankone}0.268$\pm$0.161 & \cellcolor{rankone}4.1$\pm$7.5 \\
\midrule
\textcolor{gray}{VAE ceiling} & \textcolor{gray}{0.802$\pm$0.068} & \textcolor{gray}{1.7$\pm$1.4} & \textcolor{gray}{0.877$\pm$0.061} & \textcolor{gray}{0.2$\pm$0.2} & \textcolor{gray}{0.872$\pm$0.081} & \textcolor{gray}{0.7$\pm$0.9} \\
\bottomrule
\end{tabular*}

\end{table}

\begin{figure}[t]
  \centering
  \includegraphics{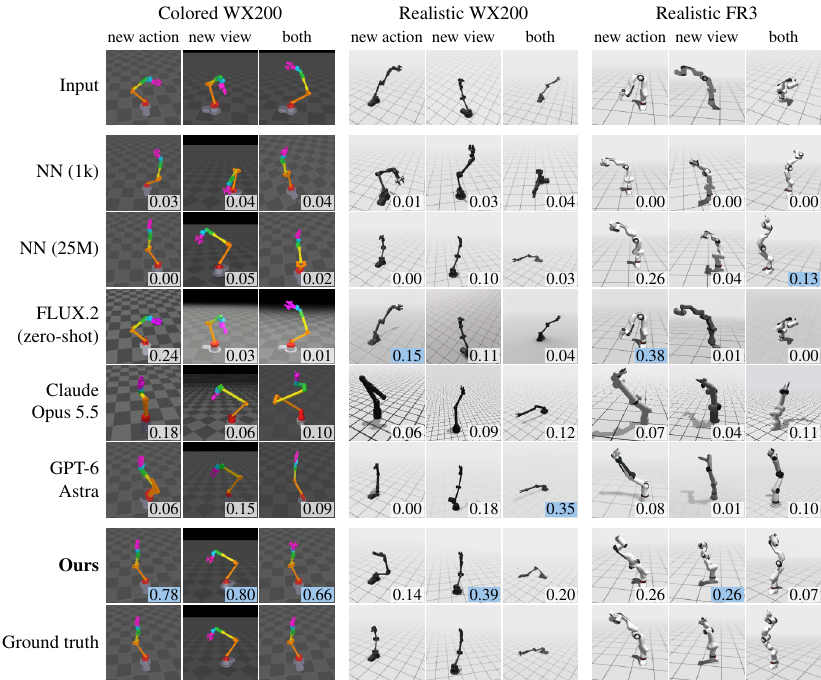}
  \caption{\textbf{Predictions on the test set}: the first test pair of each mode. Corner number: body IoU; blue: best per pair.}
  \label{fig:gallery}
\end{figure}

Table~\ref{tab:main}\footnote{\shadekey} and Fig.~\ref{fig:gallery} show the results.
Our model reaches body IoU $0.531$, $0.346$ and $0.268$ in the three settings, and places the gripper within $1.3$--$4.1$\,px.
In contrast, nearest-neighbor retrieval over 25M frames finds a frame with a nearby gripper ($12.4$--$17.3$\,px) but a different arm configuration and scores below copying the input.
FLUX.2 without fine-tuning edits the context image but leaves the arm in its context pose ($0.048$--$0.065$ body IoU, below copying).
The language models draw convincing robots, mostly with ray casters or mesh rasterisers, and on the realistic FR3 most of their programs hard-code the Franka's published kinematic parameters (Appendix~\ref{app:zeroshot}).
However, the better language model reaches $0.116$--$0.187$ body IoU, and our model leads it by $0.15$--$0.34$ (95\% bootstrap intervals above zero in every setting).
Our model also ranks first in PSNR and L1 (Table~\ref{tab:pixel}), and Appendix~\ref{app:permode} reports all 192 test pairs and each mode.
As shown in Fig.~\ref{fig:ablation}b, link IoU saturates at 20 sampling steps, our default; time per image grows linearly with the steps, and 5 steps without guidance keep 87--92\% of the link IoU at 0.041\,s per image, $7.5\times$ faster.

\subsection{From end-effector to whole body}
\label{sec:ablation}

\begin{table}[t!]
  \centering
  \caption{\textbf{Adaptation to a bent upper arm} (realistic WX200), mean $\pm$ std over 192 test pairs. Bent and original robot: test pairs of each robot, scored by the same model. Body $\Delta$: on bent-robot pairs, body IoU against the bent ground truth minus against the intact arm at the same end-effector and camera poses. Closer: share of pairs with $\Delta > 0$.}
  \label{tab:bent}
  \footnotesize
  \setlength{\tabcolsep}{2pt}
  \begin{tabular*}{\textwidth}{@{\extracolsep{\fill}}llccccc}
\toprule
& & \multicolumn{2}{c}{Bent robot} & Original robot & \multicolumn{2}{c}{Draws the bend} \\
\cmidrule(lr){3-4}\cmidrule(lr){5-5}\cmidrule(lr){6-7}
Bend & Model & Link$\uparrow$ & Grip$\downarrow$ & Link$\uparrow$ & Body $\Delta\uparrow$ & Closer$\uparrow$ \\
\midrule
\multirow{2}{*}{$30^\circ$} & No updates & 0.355$\pm$0.176 & 1.89$\pm$4.16 & 0.370$\pm$0.181 & $-$0.009$\pm$0.086 & 51\% \\
 & Adapted, 300 traj. & 0.370$\pm$0.190 & 1.89$\pm$4.83 & 0.348$\pm$0.168 & $+$0.051$\pm$0.086 & 76\% \\
\midrule
\multirow{2}{*}{$45^\circ$} & No updates & 0.341$\pm$0.162 & 1.78$\pm$2.21 & 0.370$\pm$0.181 & $-$0.043$\pm$0.102 & 35\% \\
 & Adapted, 300 traj. & 0.370$\pm$0.204 & 1.79$\pm$2.22 & 0.329$\pm$0.160 & $+$0.069$\pm$0.090 & 79\% \\
\bottomrule
\end{tabular*}

\end{table}

\begin{figure}[t!]
  \centering
  \includegraphics{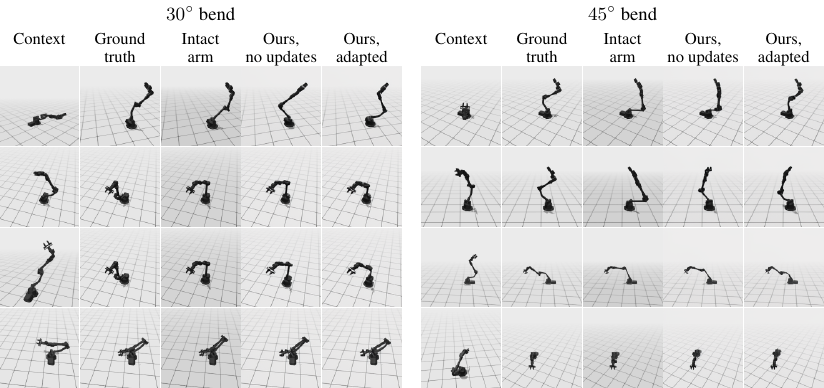}
  \caption{\textbf{A bent upper arm}, $30^\circ$ (left) and $45^\circ$ (right): the four test pairs per angle where the bend changes the arm most, picked before prediction. Intact arm: the unbent robot at the same end-effector and camera poses. Adapted: on 300 trajectories of the bent robot.}
  \label{fig:bent}
\end{figure}

\begin{figure}[t!]
  \centering
  \includegraphics{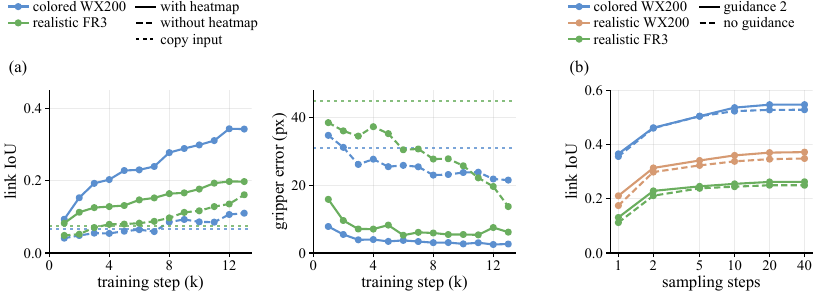}
  \caption{\textbf{Ablations on the end-effector ray heatmap and sampling steps.} (a)~Link IoU and gripper error on the 192 test pairs every 1{,}000 steps, with and without the heatmap at the same learning rate. (b)~Link IoU of the final models against sampling steps, with and without guidance.}
  \label{fig:ablation}
\end{figure}

\paragraph{A pixel-aligned action input places the arm.}
Fig.~\ref{fig:ablation}a compares the colored WX200 and realistic FR3 models with models trained without the end-effector ray heatmap at the same learning rate, where the end-effector poses still enter as condition tokens.
Without the heatmap, both models learn the arm more slowly.
At step 13{,}000, the heatmap raises link IoU from $0.110$ to $0.343$ on the colored WX200 and from $0.161$ to $0.198$ on the realistic FR3, and lowers gripper error from $21.5$ to $2.6$\,px and from $13.7$ to $6.2$\,px; the realistic FR3 model without the heatmap is still improving at 13k, and the colored WX200 one stays above 19\,px through 16k.
Yet pixel L1 changes little ($0.026$ vs $0.021$ on the colored WX200): without the heatmap the model draws a sharp robot with a misplaced arm.

\paragraph{Inverse kinematics ambiguity.}
\label{sec:ik}
A 6-DoF end-effector pose does not fix the arm of either robot: the 5-DoF WX200 reaches a pose with a few discrete configurations or with none, and the redundant 7-DoF FR3 with a continuum of them.
Targets with no solution are out of distribution (Sec.~\ref{sec:conclusion}); here we study targets with several.
Enumerating the collision-free inverse kinematics (IK) solutions of each target pose, we find that 76\% of realistic WX200 test pairs (2.6 solutions on average) and all realistic FR3 pairs have one visibly different from the ground truth (IoU $<0.8$).
To test whether this ambiguity limits the scores, we continue training each realistic model on data whose configurations follow a deterministic IK rule (the solution closest to a rest posture), with a control trained equally long on the original data.
On the realistic FR3, the rule raises link IoU by $0.041$ (95\% CI $0.031$ to $0.052$) on test targets that follow the rule and by $0.005$ on the original test targets; scored against the best-matching valid configuration, the gain is $0.019$ (Appendix~\ref{app:ik}).
In contrast, on the realistic WX200 the rule changes link IoU by $-0.003$ and $-0.022$ on the same two test sets.

\subsection{Adapting to a changed body}
\label{sec:bent}

We bend the upper arm of the realistic WX200 at mid-length by $30^\circ$ or $45^\circ$.
A bent and an intact arm that reach the same end-effector pose overlap heavily (body IoU $0.64$ at $30^\circ$), so we render each bent test pair a second time with the intact arm and measure $\Delta = \mathrm{IoU}(\hat I', \text{bent}) - \mathrm{IoU}(\hat I', \text{intact})$ on the arm links.
As a baseline, copying the context image gives $\Delta \approx 0$, with 39--40\% of pairs closer to the bent arm.

\paragraph{The model keeps the body it was trained on.}
Shown a context image of the bent robot, the model draws the arm it was trained on (Fig.~\ref{fig:bent}): $\Delta = -0.009$ (95\% CI $-0.022$ to $0.003$) at $30^\circ$ and $-0.043$ ($-0.058$ to $-0.029$) at $45^\circ$.
When the context image shows the intact arm or one of three arms bent by $45^\circ$, the predictions are closest to the intact arm in every case, and a context image showing the ground-truth arm raises body IoU by at most $0.007$ (Appendix~\ref{app:bent}).

\paragraph{Continued training adapts the model toward the new body.}
As shown in Table~\ref{tab:bent}, training for 2{,}000 steps on 300 trajectories of the bent robot raises $\Delta$ on every arm link, and 30 trajectories already raise it to $+0.035$ at $30^\circ$.
With 300 trajectories, $\Delta$ rises to $+0.051$ ($0.039$ to $0.063$) at $30^\circ$ and to $+0.069$ ($0.057$ to $0.082$) at $45^\circ$, and the share of pairs closer to the bent arm to 76\% and 79\%.
Of the gap in $\Delta$ between the model without updates and the VAE reconstruction of the bent ground truth, adaptation closes 29\% at $45^\circ$ and a bent context image alone 2\%.
Adaptation also takes over $30\times$ less compute than training anew on the bent robot (Sec.~\ref{sec:compute}).

\section{Limitations and Conclusion}
\label{sec:conclusion}

\begin{figure}[t]
  \centering
  \includegraphics{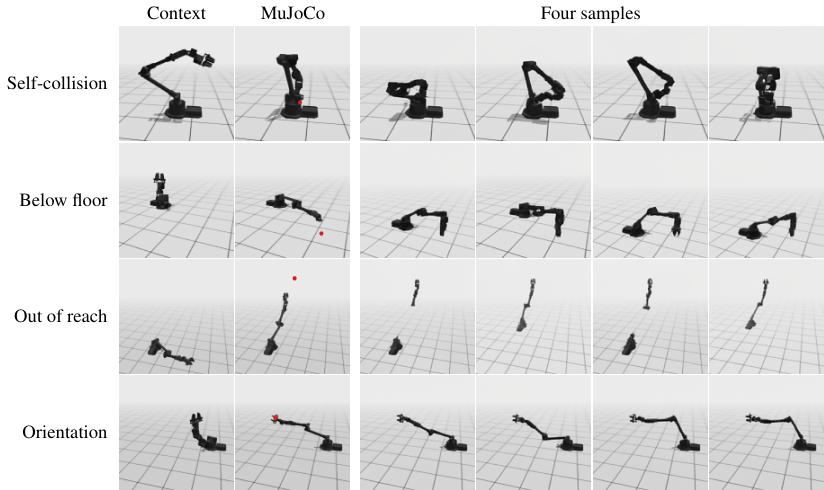}
  \caption{\textbf{Out-of-distribution action inputs} (realistic WX200, first pair of each category). MuJoCo: the colliding pose (top two rows) or the closest reachable pose (bottom two rows); red dot: requested end-effector position.}
  \label{fig:ood}
\end{figure}

\paragraph{Limitations.}
Our study has four main limitations.
(1)~All results are in simulation at $128\times128$ pixels, with one robot per model and no objects or contact; real-robot images, with estimated camera and end-effector poses, are future work.
(2)~Per image, the model is slower than its simulator (0.31\,s vs 3.5\,ms), although as a neural network it runs batched on GPUs and is differentiable in its action and camera inputs; it also has no notion of time, so chaining its predictions drifts.
(3)~The body is a soft constraint and the end-effector input a hard one: for an unreachable target the model stretches or detaches the arm (Fig.~\ref{fig:ood}, Appendix~\ref{app:ood}), and adaptation to a bent arm is partial, with little effect from the context image alone.
(4)~Ground truth contains one of several valid arm configurations, which penalises valid alternatives of every method, mostly on the realistic FR3.
Further directions are higher resolutions, a notion of time, and learning a deterministic IK rule together with the body, such as the configuration a planner reaches from a rest posture, which our clean-IK rule already shows to help on the redundant FR3 (Sec.~\ref{sec:ik}).

\paragraph{Conclusion.}
In this paper, we studied third-person, whole-body robot simulation with a pre-trained image generator. Fine-tuned on image pairs streamed from a simulator, the model renders the whole robot at a commanded end-effector pose from a new camera pose and needs no URDF, joint angles or simulator at test time.
On three robot settings, it outperforms nearest-neighbor retrieval from a database as large as its training data and language models that write rendering code.
The pixel-aligned end-effector ray heatmap places the arm, and per-link IoU measures this placement, which pixel metrics such as PSNR barely reflect because the robot covers 2--5\% of the image.
The model keeps the body it was trained on: shown a bent arm, it draws its original arm. However, brief continued training on the bent robot adapts it toward the new body.
Since views from known cameras carry the robot's geometry, we expect the approach to extend to scenes with objects or several robots in future work.

\bibliography{refs_paper}

\clearpage
\appendix
\section{General-Purpose Baselines}
\label{app:zeroshot}
\paragraph{Queries.}
Each language-model query contains the context image (upsampled to 512\,px), the context and target end-effector poses (position, rotation matrix, quaternion), both cameras (position, forward, up, field of view), the pixel positions of both end-effectors, and the coordinate conventions; no robot name or asset is given.
The returned Python program writes the $128\times128$ target image in a sandbox with numpy, scipy, Pillow and matplotlib, no network access, a 60\,s limit and one retry after an error.
Both models ran through Amazon Bedrock with high reasoning effort and 32k output tokens; every query returned an image (2 retries for Opus, 14 for Astra), at \$0.22 (Opus) and \$0.81 (Astra) per query and 2.5--5.4 minutes.
FLUX.2 (zero-shot) uses \texttt{Flux2KleinPipeline} with the context image as reference and an editing instruction followed by the same query as text, with the text encoder's maximum length raised to 1{,}344 tokens, 50 steps and guidance scale 4.
With the prompt ``the same robot'', FLUX.2 returns the context image nearly unchanged (body IoU within $0.007$ of copying).

\paragraph{Programs.}
Table~\ref{tab:llmcode} classifies the 288 programs, labelled by a rule-based classifier and checked by hand on 119 programs (97\% agreement on programs read before the rules were fixed).
Opus writes ray casters for capsules, spheres and boxes (85\% of its programs) or signed distance fields (10\%), and Astra writes triangle-mesh rasterisers (94\%); 13 other programs draw in 2-D, warp the input image or use another method, and none uses voxels.
Opus's programs average 5.9k characters and Astra's 13.3k.
On the realistic FR3, 44 of 48 Opus programs and 40 of 48 Astra programs hard-code the Franka's published kinematic parameters.
On the realistic WX200, Opus names a robot in 20 of 48 programs, the correct one in 8, and otherwise uses parameters of another arm such as a ViperX-300; Astra hard-codes WidowX-250 dimensions.
On the colored WX200, 17 Astra programs cut the colored links out of the context image and paste them at new positions as rigid pieces.

\begin{table}[h]
  \centering
  \caption{\textbf{Language-model programs}: count and share of the 48 programs per setting. Rendering and arm configuration each split the programs; the last block counts programs with each property. Hard-coded kinematics: two or more published constants of a real robot, or one with its name.}
  \label{tab:llmcode}
  \footnotesize
  \setlength{\tabcolsep}{2pt}
  \begin{tabular*}{\textwidth}{@{}l@{\extracolsep{\fill}}>{\raggedleft\arraybackslash}p{0.42cm}@{\extracolsep{0pt}\,}>{\raggedright\arraybackslash}p{0.78cm}@{\extracolsep{\fill}}>{\raggedleft\arraybackslash}p{0.42cm}@{\extracolsep{0pt}\,}>{\raggedright\arraybackslash}p{0.78cm}@{\extracolsep{\fill}}>{\raggedleft\arraybackslash}p{0.42cm}@{\extracolsep{0pt}\,}>{\raggedright\arraybackslash}p{0.78cm}@{\extracolsep{\fill}}>{\raggedleft\arraybackslash}p{0.42cm}@{\extracolsep{0pt}\,}>{\raggedright\arraybackslash}p{0.78cm}@{\extracolsep{\fill}}>{\raggedleft\arraybackslash}p{0.42cm}@{\extracolsep{0pt}\,}>{\raggedright\arraybackslash}p{0.78cm}@{\extracolsep{\fill}}>{\raggedleft\arraybackslash}p{0.42cm}@{\extracolsep{0pt}\,}>{\raggedright\arraybackslash}p{0.78cm}@{\extracolsep{\fill}}}
\toprule
 & \multicolumn{6}{c}{Claude Opus 5.5} & \multicolumn{6}{c}{GPT-6 Astra} \\
\cmidrule(lr){2-7}\cmidrule(lr){8-13}
Programs (of 48) & \multicolumn{2}{c}{\shortstack{Colored\\WX200}} & \multicolumn{2}{c}{\shortstack{Realistic\\WX200}} & \multicolumn{2}{c}{\shortstack{Realistic\\FR3}} & \multicolumn{2}{c}{\shortstack{Colored\\WX200}} & \multicolumn{2}{c}{\shortstack{Realistic\\WX200}} & \multicolumn{2}{c}{\shortstack{Realistic\\FR3}} \\
\midrule
\multicolumn{13}{l}{\emph{Rendering}} \\
Primitives (capsules, spheres, boxes) & 41 & (86\%) & 39 & (81\%) & 43 & (90\%) & 0 & (0\%) & 1 & (2\%) & 0 & (0\%) \\
Triangle meshes & 1 & (2\%) & 0 & (0\%) & 0 & (0\%) & 45 & (94\%) & 44 & (92\%) & 47 & (98\%) \\
Signed distance fields & 2 & (4\%) & 7 & (15\%) & 5 & (10\%) & 0 & (0\%) & 0 & (0\%) & 0 & (0\%) \\
2-D drawing, image warp or other & 4 & (8\%) & 2 & (4\%) & 0 & (0\%) & 3 & (6\%) & 3 & (6\%) & 1 & (2\%) \\
\midrule
\multicolumn{13}{l}{\emph{Arm configuration}} \\
Numerical IK & 8 & (17\%) & 7 & (15\%) & 44 & (92\%) & 7 & (14\%) & 3 & (6\%) & 43 & (90\%) \\
Closed-form IK & 24 & (50\%) & 26 & (54\%) & 2 & (4\%) & 33 & (69\%) & 43 & (90\%) & 2 & (4\%) \\
No IK or other & 16 & (33\%) & 15 & (31\%) & 2 & (4\%) & 8 & (17\%) & 2 & (4\%) & 3 & (6\%) \\
\midrule
\multicolumn{13}{l}{\emph{Prior knowledge and input use}} \\
Names the correct robot & 0 & (0\%) & 8 & (17\%) & 43 & (90\%) & 0 & (0\%) & 0 & (0\%) & 10 & (21\%) \\
Hard-codes published kinematics & 3 & (6\%) & 11 & (23\%) & 44 & (92\%) & 2 & (4\%) & 9 & (19\%) & 40 & (83\%) \\
Reads context-image pixels & 13 & (27\%) & 4 & (8\%) & 4 & (8\%) & 46 & (96\%) & 39 & (81\%) & 29 & (60\%) \\
\bottomrule
\end{tabular*}

\end{table}

\begin{figure}[h]
  \centering
  \includegraphics{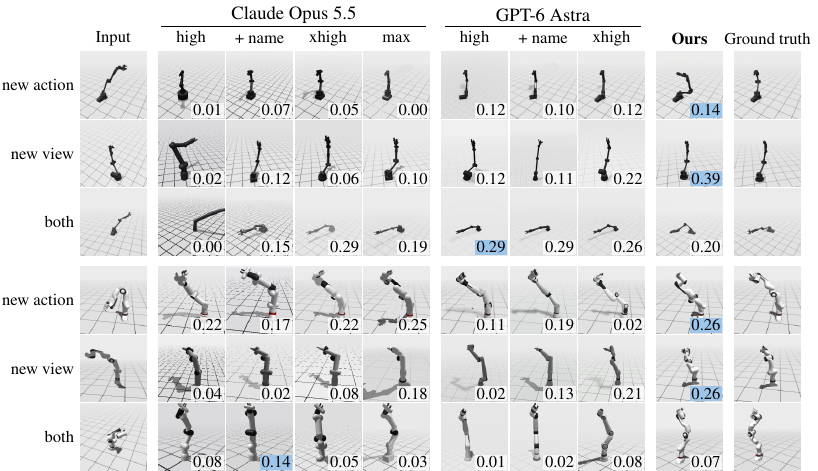}
  \caption{\textbf{Robot name and reasoning effort}, one sample per query through the Claude Code and Codex command-line tools, on the gallery pairs of Fig.~\ref{fig:gallery} (top three rows: realistic WX200; bottom three: realistic FR3). high: the image-only query of Table~\ref{tab:main}; + name: the robot named in the prompt; xhigh, max: higher effort without the name. Corner: body IoU; blue: best per pair. Mean body IoU over the six pairs: Opus 0.06, 0.11, 0.13, 0.12; Astra 0.11, 0.14, 0.15; ours 0.22.}
  \label{fig:probe}
\end{figure}

\paragraph{Robot name and reasoning effort.}
All language-model results in the main text use the query above, which shows the robot only through the context image, at high reasoning effort.
On the six gallery pairs of the realistic settings we probe two changes, with one sample per query (Fig.~\ref{fig:probe}): naming the robot in the conventions (``The robot is the Trossen Robotics WidowX 200 (WX200)'' or ``the Franka Research 3 (FR3)'', followed by an invitation to use what the model knows about it), and raising the effort to xhigh or, for Opus, max.
These queries ran through the Claude Code and Codex command-line tools with the same prompts, image, sandbox and retry; Astra writes about 40\% fewer output tokens through Codex than through Bedrock at the same setting, so each change is compared with an unnamed, high-effort query through the same tools.
Naming the robot raises mean body IoU from 0.06 to 0.11 for Opus and from 0.11 to 0.14 for Astra; effort xhigh raises it to 0.13 and 0.15, at 13--20 and 7--11 minutes per query instead of 1--7.
Opus at max effort reaches 0.12, writing 230k--340k output tokens over 36--54 minutes per query; with a 64k-token output limit, every max query ended before a program was written.
Repeating the unnamed, high-effort Opus query changes a pair's body IoU by up to 0.15, and every setting stays below ours on these pairs (0.22).
On the realistic FR3, most image-only programs already recognise the Franka and hard-code its published kinematics (Table~\ref{tab:llmcode}), and naming it adds little.

\section{IK Ambiguity}
\label{app:ik}
This section details the study of Sec.~\ref{sec:ik}.
For each test pair we enumerate the collision-free IK solutions of the target end-effector pose: all discrete branches for the WX200 (2.6 per pair) and 28 samples of the continuous self-motion for the FR3.
The clean-IK model continues training on data whose configurations follow the rule (the solution closest to a rest posture) at half the original learning rate; the control continues equally long on the original data.
Both runs use early stopping with the same rule.
Table~\ref{tab:ik} scores both on targets drawn either way, against the ground truth and against the best-matching valid configuration.
On the FR3 the rule's own solution is always a candidate while the self-motion is sampled, so the gain against the best valid configuration is an upper bound.

\begin{table}[h]
  \centering
  \caption{\textbf{Clean-IK model minus control}: link IoU difference with 95\% bootstrap interval over 192 test pairs (4 seeds). Best valid: per pair, the best link IoU over all candidate configurations.}
  \label{tab:ik}
  \footnotesize
  \setlength{\tabcolsep}{2pt}
  \begin{tabular*}{\textwidth}{@{\extracolsep{\fill}}llcc}
    \toprule
    Robot & Targets & vs ground truth & vs best valid \\
    \midrule
    Realistic WX200 & sampled & $-$0.022 {\scriptsize[$-$0.033, $-$0.011]} & $-$0.010 {\scriptsize[$-$0.022, $+$0.001]} \\
                    & clean rule & $-$0.003 {\scriptsize[$-$0.015, $+$0.009]} & $-$0.010 {\scriptsize[$-$0.021, $+$0.000]} \\
    Realistic FR3   & sampled & $+$0.005 {\scriptsize[$-$0.004, $+$0.013]} & $+$0.014 {\scriptsize[$+$0.006, $+$0.022]} \\
                    & clean rule & $\mathbf{+}$\textbf{0.041} {\scriptsize[$+$0.031, $+$0.052]} & $+$0.019 {\scriptsize[$+$0.010, $+$0.027]} \\
    \bottomrule
  \end{tabular*}
\end{table}

\section{Bent Arm}
\label{app:bent}
The intact arm in Table~\ref{tab:bent} is the IK solution most similar to the bent ground truth.
Against the best of all intact solutions (2.3 per pair), $\Delta$ is $-0.056$ without updates and $+0.013$ after adaptation at $30^\circ$, and $-0.081$ and $+0.041$ at $45^\circ$.
Without updates, $\Delta$ is $+0.010$, $-0.009$ and $-0.043$ for bends of $15^\circ$, $30^\circ$ and $45^\circ$.
At $45^\circ$, adaptation raises link IoU on the bent robot from $0.341$ to $0.370$, and $\Delta$ on the bent link from $-0.073$ to $+0.100$ and on the next link from $-0.082$ to $+0.133$.

\paragraph{Context swap.}
For 183 test pairs we render the context image with four arms (intact, upper arm bent $+45^\circ$ or $-45^\circ$, forearm bent $+45^\circ$), query the model without updates with each, and score the predictions against all four ground truths (Table~\ref{tab:ctxswap}).
The $-45^\circ$ arm is close to the intact one (body IoU $0.81$ between the ground truths, against $0.54$ for $+45^\circ$).

\paragraph{Gap closed.}
For a prediction $X$ with body-link difference $d_X$, the fraction of the gap closed is $(d_X - d_0)/(d_1 - d_0)$, where $d_0 = -0.053$ is the model without updates shown the intact arm and $d_1 = +0.376$ the VAE reconstruction of the bent ground truth.
Over 187 pairs, the bent context image reaches $0.02$ ($0.01$ to $0.04$) and the adapted model $0.29$ ($0.24$ to $0.33$).

\begin{table}[h]
  \centering
  \caption{\textbf{Context swap} ($45^\circ$ bends, model without updates, 183 pairs): body IoU, mean $\pm$ std. Rows: context arm; columns: ground-truth arm; last row: matching context minus the mean of the other three, with 95\% bootstrap interval.}
  \label{tab:ctxswap}
  \scriptsize
  \setlength{\tabcolsep}{2pt}
  \begin{tabular*}{\textwidth}{@{\extracolsep{\fill}}lcccc}
\toprule
& \multicolumn{4}{c}{Ground-truth arm} \\
\cmidrule(lr){2-5}
Context arm & Intact & Upper arm $+45^\circ$ & Upper arm $-45^\circ$ & Forearm $+45^\circ$ \\
\midrule
Intact & \cellcolor{rankone}0.326$\pm$0.197 & 0.279$\pm$0.145 & \cellcolor{rankone}0.312$\pm$0.173 & \cellcolor{rankone}0.289$\pm$0.166 \\
Upper arm $+45^\circ$ & 0.319$\pm$0.197 & \cellcolor{rankone}0.284$\pm$0.149 & 0.306$\pm$0.173 & 0.286$\pm$0.167 \\
Upper arm $-45^\circ$ & 0.320$\pm$0.192 & 0.275$\pm$0.144 & 0.310$\pm$0.171 & 0.284$\pm$0.162 \\
Forearm $+45^\circ$ & 0.319$\pm$0.193 & 0.278$\pm$0.148 & 0.311$\pm$0.174 & 0.287$\pm$0.166 \\
\midrule
Own $-$ other & $+$0.006 [$+$0.002, $+$0.011] & $+$0.007 [$+$0.003, $+$0.011] & $+$0.000 [$-$0.005, $+$0.005] & $+$0.001 [$-$0.004, $+$0.005] \\
\bottomrule
\end{tabular*}

\end{table}

\section{Out-of-Distribution Action Inputs}
\label{app:ood}
We query each model with target end-effector poses its data engine never produces: 16 new-action pairs per category with a normal context image, 4 sampling seeds each.
\emph{Self-collision} and \emph{below floor} targets are reachable only with interpenetrating links ($\ge 5$\,mm) or a link at least 2\,cm below the floor; \emph{out of reach} targets lie 1.2--1.6$\times$ further from the shoulder; \emph{orientation} targets have a reachable position and an orientation at least $20^\circ$ from any reachable one.
Out of reach, the gripper lands $5$--$16$\,px from the requested point, against $16$--$22$\,px for the closest reachable pose, and the four samples agree as much as on valid inputs (Table~\ref{tab:ood}); the arm is stretched or detached from its base in about a third of the realistic WX200 samples (Fig.~\ref{fig:ood}).
For collision targets the model draws a collision-free arm near the point, and its samples agree less ($0.16$--$0.39$ against $0.29$--$0.66$ in distribution).
For orientation targets it draws a plausible arm with the gripper in place.

\begin{table}[H]
  \centering
  \caption{\textbf{Out-of-distribution action inputs}, mean $\pm$ std over 16 pairs. Ours, MuJoCo: gripper-centroid distance in pixels to the requested end-effector pixel; agreement: mean pairwise body IoU of the four samples; --: the redundant FR3 reaches nearly every orientation at a reachable position.}
  \label{tab:ood}
  \footnotesize
  \setlength{\tabcolsep}{4pt}
  \resizebox{\textwidth}{!}{%
\begin{tabular}{@{}lccccccccc@{}}
\toprule
 & \multicolumn{3}{c}{Colored WX200} & \multicolumn{3}{c}{Realistic WX200} & \multicolumn{3}{c}{Realistic FR3} \\
\cmidrule(lr){2-4}\cmidrule(lr){5-7}\cmidrule(lr){8-10}
Target & Ours$\downarrow$ & MuJoCo & Agreement & Ours$\downarrow$ & MuJoCo & Agreement & Ours$\downarrow$ & MuJoCo & Agreement \\
\midrule
In distribution & 9.7$\pm$4.0 & 6.8$\pm$3.6 & 0.66$\pm$0.15 & 7.5$\pm$3.8 & 6.8$\pm$3.6 & 0.56$\pm$0.16 & 5.0$\pm$2.0 & 4.2$\pm$2.1 & 0.29$\pm$0.10 \\
\midrule
Self-collision & 6.8$\pm$2.5 & 7.7$\pm$2.3 & 0.35$\pm$0.14 & 10.3$\pm$3.2 & 7.7$\pm$2.3 & 0.25$\pm$0.09 & 8.0$\pm$3.6 & 5.3$\pm$2.8 & 0.16$\pm$0.06 \\
Below floor & 12.0$\pm$9.0 & 5.2$\pm$2.1 & 0.39$\pm$0.21 & 7.2$\pm$2.9 & 5.2$\pm$2.1 & 0.29$\pm$0.18 & 9.2$\pm$7.3 & 5.9$\pm$2.5 & 0.21$\pm$0.08 \\
Out of reach & 16.4$\pm$13.8 & 21.7$\pm$7.1 & 0.48$\pm$0.15 & 8.3$\pm$3.9 & 21.7$\pm$7.1 & 0.53$\pm$0.17 & 5.2$\pm$2.2 & 16.3$\pm$6.1 & 0.40$\pm$0.10 \\
Orientation & 9.4$\pm$3.8 & 7.7$\pm$2.7 & 0.62$\pm$0.19 & 7.4$\pm$4.7 & 7.7$\pm$2.7 & 0.54$\pm$0.20 & -- & -- & -- \\
\bottomrule
\end{tabular}%
}

\end{table}

\clearpage
\section{Additional Results}
\label{app:permode}
\label{app:pixel}
Tables~\ref{tab:pixel}--\ref{tab:permode} extend Table~\ref{tab:main}: pixel metrics on its 48 pairs, all metrics on all 192 test pairs, and body IoU per mode.
PSNR and L1 rank our model first in every setting, with margins of $0.6$--$6.7$\,dB PSNR over the next method; the robot covers 2--5\% of the pixels, so pixel metrics mostly score the background.
On all 192 pairs, new actions are easier than new views on the realistic robots (body IoU $0.478$ vs $0.254$ on the realistic WX200), and on the realistic FR3 the wrist and gripper score like the body ($0.266$ vs $0.259$).

\begin{table}[H]
  \centering
  \caption{\textbf{Pixel metrics} on the 48 test pairs of Table~\ref{tab:main}, mean $\pm$ std over pairs; ours: 4 sampling seeds. PSNR in dB and L1 on $128\times128$ images with values in $[0,1]$.}
  \label{tab:pixel}
  \footnotesize
  \setlength{\tabcolsep}{2pt}
  \begin{tabular*}{\textwidth}{@{\extracolsep{\fill}}lcccccc}
\toprule
 & \multicolumn{2}{c}{Colored WX200} & \multicolumn{2}{c}{Realistic WX200} & \multicolumn{2}{c}{Realistic FR3} \\
\cmidrule(lr){2-3}\cmidrule(lr){4-5}\cmidrule(lr){6-7}
Method & PSNR$\uparrow$ & L1$\downarrow$ & PSNR$\uparrow$ & L1$\downarrow$ & PSNR$\uparrow$ & L1$\downarrow$ \\
\midrule
Random frame & 18.5$\pm$3.0 & 0.080$\pm$0.039 & 14.9$\pm$1.1 & 0.076$\pm$0.013 & 16.3$\pm$1.3 & 0.078$\pm$0.015 \\
Copy input & 20.3$\pm$3.8 & 0.059$\pm$0.046 & \cellcolor{rankthree}17.0$\pm$3.6 & \cellcolor{rankthree}0.055$\pm$0.029 & \cellcolor{rankthree}17.5$\pm$2.7 & \cellcolor{rankthree}0.062$\pm$0.032 \\
NN (1k) & 20.7$\pm$2.1 & 0.048$\pm$0.014 & 15.2$\pm$1.7 & 0.071$\pm$0.016 & 16.3$\pm$1.3 & 0.073$\pm$0.013 \\
NN (25M) & \cellcolor{rankthree}21.8$\pm$2.1 & \cellcolor{ranktwo}0.037$\pm$0.009 & 15.8$\pm$2.0 & 0.063$\pm$0.018 & 16.8$\pm$1.7 & 0.066$\pm$0.017 \\
FLUX.2 (zero-shot) & 15.4$\pm$3.7 & 0.155$\pm$0.077 & 14.8$\pm$1.7 & 0.114$\pm$0.035 & 15.3$\pm$1.7 & 0.122$\pm$0.026 \\
Claude Opus 5.5 & 20.9$\pm$2.7 & 0.061$\pm$0.026 & 14.7$\pm$4.1 & 0.109$\pm$0.045 & 15.1$\pm$1.3 & 0.106$\pm$0.023 \\
GPT-6 Astra & \cellcolor{ranktwo}22.9$\pm$3.4 & \cellcolor{rankthree}0.040$\pm$0.026 & \cellcolor{ranktwo}18.4$\pm$3.4 & \cellcolor{ranktwo}0.048$\pm$0.021 & \cellcolor{ranktwo}18.8$\pm$2.3 & \cellcolor{ranktwo}0.054$\pm$0.022 \\
Ours & \cellcolor{rankone}29.6$\pm$3.6 & \cellcolor{rankone}0.011$\pm$0.005 & \cellcolor{rankone}20.7$\pm$4.2 & \cellcolor{rankone}0.037$\pm$0.017 & \cellcolor{rankone}19.4$\pm$2.4 & \cellcolor{rankone}0.043$\pm$0.018 \\
\midrule
\textcolor{gray}{VAE ceiling} & \textcolor{gray}{38.0$\pm$1.0} & \textcolor{gray}{0.004$\pm$0.001} & \textcolor{gray}{37.6$\pm$0.8} & \textcolor{gray}{0.009$\pm$0.001} & \textcolor{gray}{36.6$\pm$0.7} & \textcolor{gray}{0.009$\pm$0.001} \\
\bottomrule
\end{tabular*}

\end{table}

\begin{table}[H]
  \centering
  \caption{\textbf{Main results on all 192 test pairs} per setting, mean $\pm$ std over pairs; ours: 4 sampling seeds. L1: pixel error.}
  \label{tab:main192}
  \footnotesize
  \setlength{\tabcolsep}{4pt}
  \resizebox{\textwidth}{!}{%
\begin{tabular}{@{}lccccccccc@{}}
\toprule
 & \multicolumn{3}{c}{Colored WX200} & \multicolumn{3}{c}{Realistic WX200} & \multicolumn{3}{c}{Realistic FR3} \\
\cmidrule(lr){2-4}\cmidrule(lr){5-7}\cmidrule(lr){8-10}
Method & Body$\uparrow$ & Grip$\downarrow$ & L1$\downarrow$ & Body$\uparrow$ & Grip$\downarrow$ & L1$\downarrow$ & Body$\uparrow$ & Grip$\downarrow$ & L1$\downarrow$ \\
\midrule
Random frame & 0.020$\pm$0.033 & 34.3$\pm$16.9 & 0.077$\pm$0.039 & 0.018$\pm$0.036 & 38.4$\pm$18.5 & 0.076$\pm$0.015 & 0.030$\pm$0.041 & 46.7$\pm$21.3 & 0.079$\pm$0.014 \\
Copy input & 0.092$\pm$0.122 & 31.0$\pm$18.3 & 0.056$\pm$0.046 & 0.098$\pm$0.140 & 33.4$\pm$20.1 & 0.057$\pm$0.029 & 0.101$\pm$0.133 & 44.9$\pm$23.3 & 0.064$\pm$0.032 \\
NN (1k) & 0.024$\pm$0.036 & 22.4$\pm$12.3 & 0.047$\pm$0.015 & 0.028$\pm$0.043 & 21.7$\pm$13.4 & 0.070$\pm$0.016 & 0.038$\pm$0.048 & 28.7$\pm$17.3 & 0.072$\pm$0.015 \\
NN (25M) & 0.043$\pm$0.066 & 13.5$\pm$7.8 & 0.038$\pm$0.009 & 0.035$\pm$0.061 & 14.0$\pm$7.9 & 0.064$\pm$0.018 & 0.068$\pm$0.086 & 19.5$\pm$16.9 & 0.067$\pm$0.017 \\
FLUX.2 (zero-shot) & 0.072$\pm$0.103 & 31.4$\pm$18.6 & 0.162$\pm$0.073 & 0.062$\pm$0.089 & 33.3$\pm$17.7 & 0.113$\pm$0.033 & 0.043$\pm$0.086 & 42.0$\pm$21.4 & 0.123$\pm$0.027 \\
Ours & \cellcolor{rankone}0.518$\pm$0.185 & \cellcolor{rankone}2.0$\pm$1.9 & \cellcolor{rankone}0.011$\pm$0.006 & \cellcolor{rankone}0.325$\pm$0.213 & \cellcolor{rankone}1.5$\pm$2.5 & \cellcolor{rankone}0.039$\pm$0.019 & \cellcolor{rankone}0.259$\pm$0.157 & \cellcolor{rankone}5.0$\pm$10.5 & \cellcolor{rankone}0.045$\pm$0.018 \\
\midrule
\textcolor{gray}{VAE ceiling} & \textcolor{gray}{0.802$\pm$0.075} & \textcolor{gray}{1.4$\pm$1.3} & \textcolor{gray}{0.004$\pm$0.000} & \textcolor{gray}{0.878$\pm$0.071} & \textcolor{gray}{0.3$\pm$0.3} & \textcolor{gray}{0.009$\pm$0.001} & \textcolor{gray}{0.882$\pm$0.071} & \textcolor{gray}{1.2$\pm$5.0} & \textcolor{gray}{0.009$\pm$0.001} \\
\bottomrule
\end{tabular}%
}

\end{table}

\begin{table}[H]
  \centering
  \caption{\textbf{Body IoU per mode} on all 192 test pairs (64 per mode), mean $\pm$ std over pairs.}
  \label{tab:permode}
  \footnotesize
  \setlength{\tabcolsep}{4pt}
  \resizebox{\textwidth}{!}{%
\begin{tabular}{@{}lccccccccc@{}}
\toprule
 & \multicolumn{3}{c}{Colored WX200} & \multicolumn{3}{c}{Realistic WX200} & \multicolumn{3}{c}{Realistic FR3} \\
\cmidrule(lr){2-4}\cmidrule(lr){5-7}\cmidrule(lr){8-10}
Method & new action & new view & both & new action & new view & both & new action & new view & both \\
\midrule
Copy input & 0.235$\pm$0.109 & 0.021$\pm$0.036 & 0.019$\pm$0.029 & 0.248$\pm$0.149 & 0.026$\pm$0.040 & 0.021$\pm$0.043 & 0.243$\pm$0.140 & 0.033$\pm$0.046 & 0.028$\pm$0.037 \\
NN (25M) & 0.051$\pm$0.083 & 0.040$\pm$0.057 & 0.038$\pm$0.052 & 0.043$\pm$0.066 & 0.025$\pm$0.041 & 0.037$\pm$0.070 & 0.084$\pm$0.090 & 0.060$\pm$0.088 & 0.060$\pm$0.079 \\
Ours & \cellcolor{rankone}0.583$\pm$0.167 & \cellcolor{rankone}0.505$\pm$0.171 & \cellcolor{rankone}0.467$\pm$0.195 & \cellcolor{rankone}0.478$\pm$0.195 & \cellcolor{rankone}0.254$\pm$0.180 & \cellcolor{rankone}0.242$\pm$0.174 & \cellcolor{rankone}0.299$\pm$0.172 & \cellcolor{rankone}0.247$\pm$0.147 & \cellcolor{rankone}0.232$\pm$0.140 \\
\bottomrule
\end{tabular}%
}

\end{table}

\end{document}